\pdfoutput=1

\documentclass[11pt]{article}

\usepackage{acl}

\usepackage{times}
\usepackage{latexsym}

\usepackage[T1]{fontenc}

\usepackage[utf8]{inputenc}

\usepackage{microtype}

\usepackage{amsfonts}
\usepackage{bbm}
\usepackage{times}
\usepackage{latexsym}
\usepackage{booktabs} 
\usepackage{graphicx}
\usepackage{placeins}
\usepackage{float}
\usepackage{subfigure}
\usepackage{amssymb}
\usepackage{amsmath}
\usepackage{multirow}
 \usepackage{longtable}

\title{Nearest Neighbor Knowledge Distillation for Neural Machine Translation}

\author{Zhixian Yang \and Renliang Sun \and Xiaojun Wan \\
  Wangxuan Institute of Computer Technology, Peking University \\
  Center for Data Science, Peking University \\
  The MOE Key Laboratory of Computational Linguistics, Peking University \\
  \texttt{yangzhixian@stu.pku.edu.cn} \\
  \texttt{sunrenliangpku@gmail.com} \\
  \texttt{wanxiaojun@pku.edu.cn} \\}
  
\begin{document}
\maketitle
\begin{abstract}
k-nearest-neighbor machine translation ($k$NN-MT), proposed by \citet{knnmt}, has achieved many state-of-the-art results in machine translation tasks. Although effective, $k$NN-MT requires conducting $k$NN searches through the large datastore for each decoding step during inference, prohibitively increasing the decoding cost and thus leading to the difficulty for the deployment in real-world applications. In this paper, we propose to move the time-consuming $k$NN search forward to the preprocessing phase, and then introduce $k$ Nearest Neighbor Knowledge Distillation ($k$NN-KD) that trains the base NMT model to directly learn the knowledge of $k$NN. Distilling knowledge retrieved by $k$NN can encourage the NMT model to take more reasonable target tokens into consideration, thus addressing the overcorrection problem. Extensive experimental results show that, the proposed method achieves consistent improvement over the state-of-the-art baselines including $k$NN-MT, while maintaining the same training and decoding speed as the standard NMT model.\footnote{Our code is available at \url{https://github.com/FadedCosine/kNN-KD}}

\end{abstract}

\section{Introduction}
\label{sec-introduction}
Neural machine translation (NMT) has shown impressive progress with the prevalence of deep neural networks
\cite{transformer, or-nmt, BERT-KD}. Recently, \citet{knnmt} have proposed $k$-nearest-neighbor machine translation ($k$NN-MT) that first stores context representations and target tokens into a large datastore, and then retrieves $k$ possible target tokens by conducting nearest search from the datastore to help with the final next-token decision. The results show that $k$NN-MT can significantly improve the performance over the base NMT model.

Despite the outstanding performance, $k$NN-MT will drastically increase the testing runtime since each decoding step needs to conduct $k$NN searches \cite{fast-knnmt}. How to speed up the decoding of $k$NN-MT without degrading performance still remains an open question. Several recent works \cite{fast-knnmt, faster-knnmt} introduce some elaborate strategies to compress the datastore in which $k$NN searches are conducted, thus improving decoding efficiency to some extent. However, we argue that, where there is a time-consuming $k$NN search in the decoding phase, there is the prohibitive decoding cost, which makes it hard to be deployed on real-world applications.

In order to address the aforementioned issue more thoroughly, it is necessary to figure out why $k$NN-MT performs so well. The standard NMT models are typically trained with cross-entropy (CE)
loss with teacher forcing technique, which requires a strict word-by-word matching between the model prediction and the ground-truth. In natural language, a sentence usually has more than one expression. However, even when the model predicts a word that is reasonable but deviates from the ground-truth, the CE loss will treat it as an error and punish the model. This phenomenon is called \textit{overcorrection} \cite{or-nmt}, which often seriously harms the generalizability of NMT models. We conclude that $k$NN-MT can alleviate the problem of overcorrection by retrieving more reasonable target tokens in the decoding phase.

One natural question can be raised: can we train the model to directly learn the knowledge of $k$NN in the training phase, thus maintaining the standard decoding process without any additional decoding cost? To answer this question, we propose $k$ Nearest Neighbor Knowledge Distillation ($k$NN-KD) to distill the knowledge of the non-parametric model, i.e., $k$NN, into the base NMT model in the training phase. In detail, we first construct the datastore and then conduct $k$NN searches immediately. These two steps can be done offline in the preprocessing phase. During training, a teacher distribution $p^T_{\mathrm{kNN}}$ can be easily computed using the pre-stored results of $k$NN searches to train the NMT model to directly learn the knowledge of $k$NN. At inference time, $k$NN searches are not required, so the decoding speed is as fast as the base NMT model.  Therefore, $k$NN-KD can achieve two desirable goals simultaneously: preventing overcorrection (effectiveness) and reducing decoding cost (efficiency).


We conduct experiments on two widely acknowledged NMT benchmarks: IWSLT'14 German-English and IWSLT'15 English-Vietnamese. Experimental results show that our $k$NN-KD maintains the same training and decoding speed as the standard NMT model, while it outperforms vanilla $k$NN-MT and all the other KD methods, and yields an improvement of $+2.14$ and $+1.51$ BLEU points over the Transformer baseline. We further verify that $k$NN-KD can be adapted to diverse domains by performing experiments on multi-domains translation datasets \cite{multi-domain-data} and achieving $2.56$ BLEU improvement over vanilla $k$NN-MT on average.

In summary, the contributions of our work are as follows:
\begin{itemize}
    \item We propose $k$NN-KD that considers the $k$NN distribution as a teacher to guide the training of the base NMT model (Section~\ref{sec-knnkd}).
    \item We prove that our proposed $k$NN-KD can help to address the overcorrection issue with theoretical analysis (Section~\ref{sec-theory}).
    \item  Quantitative and qualitative results on different translation tasks validate the effectiveness and efficiency of our method (Section~\ref{sec-experiments}). 
\end{itemize}

\section{Background}
\subsection{Neural Machine Translation}

The goal of the standard NMT model is to learn the conditional probability $p_\mathrm{MT}\left(\mathbf{y} \mid \mathbf{x}\right)$ for translating a sentence $\mathbf{x} =\left\{x_{1}, \cdots, x_{m}\right\}$ in source language to a sentence $\mathbf{y}=\left\{y_{1}, \cdots, y_{n}\right\}$ in target language. Translation is usually performed in a autoregressive manner, and its probability can be factored as $p_\mathrm{MT}\left(\mathbf{y} \mid \mathbf{x}\right) = \Pi_{i=1}^{n} p\left(y_i \mid \mathbf{x}, \mathbf{y}_{<i}\right) $.  When predicting $i$-th token in the target sentence given $(\mathbf{x}, \mathbf{y}_{<i})$ as the \textit{translation context}, the NMT model encodes the translation context into a hidden state $h_{i-1}$, and outputs a probability distribution over vocabulary $\mathcal{V}$ as follows:
\begin{equation}
\label{eq-nmt}
    p_{\mathrm{MT}}\left(y_i \mid \mathbf{x}, \mathbf{y}_{<i}\right)=\frac{\exp( \mathbf{o}_{y_i}^{\top} \mathbf{h}_{i-1}) }{\sum_{w \in \mathcal{V}} \exp( \mathbf{o}_{w}^{\top} \mathbf{h}_{i-1} )},
\end{equation}
where $\mathbf{o}_{y}$ is the output embedding for $w \in \mathcal{V}$. 

We denote the ground-truth target sentence as $\mathbf{y}^\star=\left\{y_{1}^\star, \cdots, y_{n}^\star\right\}$, and for each $y_i^\star$ in the training set, the CE loss is usually used for optimizing NMT models:

\begin{equation}
\label{eq-ce}
\mathcal{L}_{\mathrm{CE}}=-\sum_{y_i \in \mathcal{V}} \mathbbm{1}_{y_i=y_{i}^{*}} \log p_{\mathrm{MT}}\left(y_i \mid \mathbf{x}, \mathbf{y}^\star_{<i}\right),
\end{equation}
where $\mathbbm{1}$ is the indicator function, and the ground-truth target sequence $\mathbf{y}^\star_{<i}$ is used in the conditions of $p_{\mathrm{MT}}$ due to the  teacher forcing technique.

\subsection{Nearest Neighbor Machine Translation}
\label{sec-vanilla-knnmt}
$k$NN-MT applies the nearest neighbor retrieval mechanism to the decoding phase of a NMT model, which allows the model direct access to a large-scale datastore for better inference. Specifically, $k$NN-MT includes two following steps: 

\noindent{\textbf{Datastore Building}
} \quad Given a bilingual sentence pair in the training set $(\mathbf{x}, \mathbf{y}^\star) \in(\mathcal{X}, \mathcal{Y}^\star)$,  $k$NN-MT first constructs a datastore $\mathcal{D}$ as follows:
\begin{equation}
\small
    (\mathcal{K}, \mathcal{V})=\bigcup_{(\mathbf{x}, \mathbf{y}^\star) \in(\mathcal{X}, \mathcal{Y}^\star)} \left\{\left(f\left(\mathbf{x}, \mathbf{y}^\star_{<i}\right), y^\star_{i}\right), \forall y^\star_{i} \in  \mathbf{y}^\star \right\},
\end{equation} where the keys are the mapping representations of all the translation contexts in the training set using the projection $f(\cdot)$, and the values are corresponding ground-truth tokens.

\begin{figure*}[t]
	\begin{center}
		\centerline{\includegraphics[width=\textwidth]{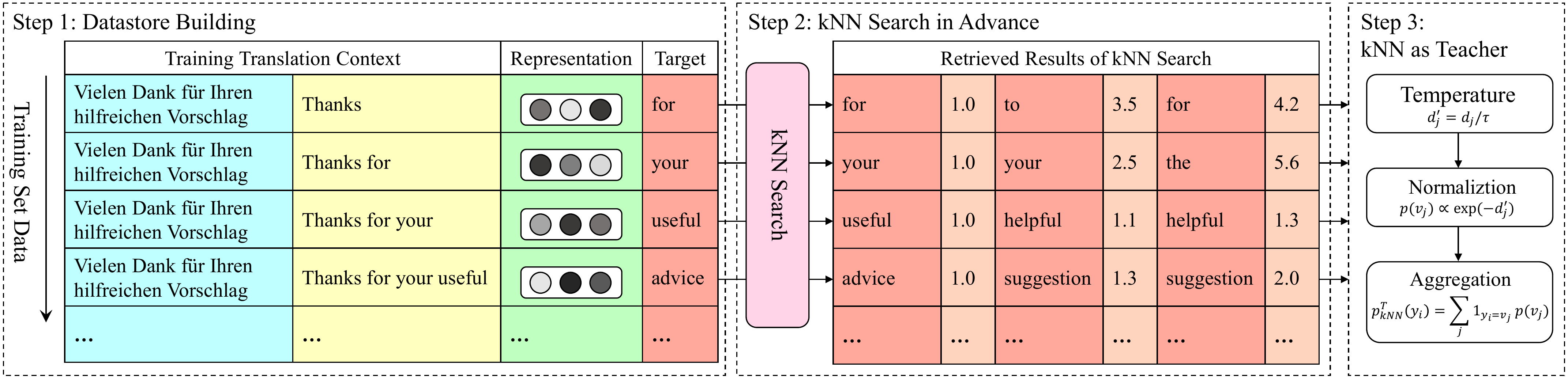}}
		\caption{Illustration of $k$NN-KD. In the preprocessing phase, we finish the datastore building in Step $1$, and conduct $k$NN search in advance in Step $2$. These two steps can be done offline before training and inference. During training, we compute the $k$NN distribution as a teacher to train the base NMT model in Step $3$. During inference, the model performs Step $4$ to decode text in the standard Seq2Seq manner, which is omitted in this figure.}
		\label{fig-knnkd}
	\end{center}
	\vskip -0.1in
\end{figure*}

\noindent{\textbf{Decoding}
} \quad During inference, $k$NN-MT aims to interpolate the base NMT model's probability in Equation~\ref{eq-nmt} with a $k$NN model. At each decoding step $i$, $k$NN-MT maps the current translation context to a representation $f\left(\mathbf{x}, \mathbf{y}_{<i}\right)$, which is used to query the datastore for $k$ nearest neighbors according to the $l_2$ distances. Denote the retrieved neighbors as $\mathcal{N}^i = \left\{\left(\mathbf{k}_{j}, v_{j}\right), j \in\{1,2, \ldots, k\}\right\} $, and then a $k$NN distribution over vocabulary $\mathcal{V}$ can be computed as:
\begin{equation}
\begin{aligned}
\label{eq-knn-prob}
&p_{\mathrm{kNN}} ( y_{i} \mid \mathbf{x}, \mathbf{y}_{<i}) \propto \\
&\sum_{\left(\mathbf{k}_{j}, v_{j} \right)\in \mathcal{N}^i } \mathbbm{1}_{y_{i}=v_{j}} \exp \left(\frac{-d\left(\mathbf{k}_{j}, f\left(\mathbf{x}, \mathbf{y}_{<i}\right) \right)}{\tau} \right),
\end{aligned}
\end{equation}
where $\tau$ is the temperature, and $d(\cdot, \cdot)$ is the $l_2$ distance function. The final probability for the next token in $k$NN-MT is the interpolation of  $p_{\mathrm{MT}}\left(y_i \mid \mathbf{x}, \mathbf{y}_{<i}\right)$ and $p_{\mathrm{kNN}} \left( y_{i}\mid\mathbf{x}, \mathbf{y}_{<i}\right)$ with a tunable weight $\lambda$:
\begin{equation}
\begin{aligned}
\label{eq-knnmt-final-prob}
   p\left(y_i \mid \mathbf{x}, \mathbf{y}_{<i}\right) &= (1 - \lambda) p_{\mathrm{MT}}\left(y_i \mid \mathbf{x}, \mathbf{y}_{<i}\right) \\
   &+ \lambda p_{\mathrm{kNN}} \left( y_{i}\mid\mathbf{x},\mathbf{y}_{<i}\right).
\end{aligned}
\end{equation}

Note that each decoding step of each beam requires a $k$NN search over the whole datastore $\mathcal{D}$, whose time complexity is $\mathcal{O}(|\mathcal{D}| B n)$ where $B$ is the beam size, and $n$ is the target length. The prohibitive decoding cost makes it hard for $k$NN-MT to be deployed on real-world applications.

\subsection{Knowledge Distillation}
\label{sec-kd}
Knowledge Distillation (KD) \cite{KD} refers to the transfer of knowledge from one neural
network T (called ``teacher model'') to another network S (called ``student model''). 

For convenience, we introduce the details of KD from the perspective of machine translation. Let $\mathbf{z} \in \mathcal{R}^{|\mathcal{V}|}$ denote the logits over $\mathcal{V}$.
Student model S outputs the probability:
\begin{equation}
p^{\mathrm{S}}\left(y_i \mid \mathbf{x}, \mathbf{y}_{<i}\right)=\frac{\exp \left(\mathbf{z}_{y_i}\right)}{\sum_{w \in \mathcal{V}} \exp \left(\mathbf{z}_{w}\right)},
\end{equation}
where $\mathbf{z}_w$ is the logit for token $w$. Correspondingly, teacher model T also predicts the probability in the same way, and a temperature factor $\tau$ can be introduced to soften the teacher’s outputs as:
\begin{equation}
p^{\mathrm{T}}\left(y_i \mid \mathbf{x}, \mathbf{y}_{<i}\right)=\frac{\exp \left(\mathbf{z}_{y_i} / \tau\right)}{\sum_{w \in \mathcal{V}} \exp \left(\mathbf{z}_{w} / \tau\right)}.
\end{equation}
When $\tau \rightarrow \infty$, $p^{\mathrm{T}}$ degenerates into a uniform distribution, and when $\tau \rightarrow 0$, $p^{\mathrm{T}}$ becomes an one-hot vector. Specifically, KD defines the objective as:
\begin{equation}
\begin{aligned}
\label{eq-single-kd}
\mathcal{L}_{\mathrm{KD}}=&-\sum_{y_i\in \mathcal{V}} p^{\mathrm{T}}\left(y_i \mid \mathbf{x}, \mathbf{y}^\star_{<i}\right) \\
& \times \log p^{\mathrm{S}}\left(y_i \mid \mathbf{x}, \mathbf{y}^\star_{<i}\right).
\end{aligned}
\end{equation}

When we apply KD to improve the performance of machine translation, student model S is usually the NMT model that will be used for testing. And then, the overall training procedure is to minimize
the summation of Equation~\ref{eq-ce} and Equation~\ref{eq-single-kd}:
\begin{equation}
\label{eq-kd-final}
\mathcal{L}=(1- \alpha)\mathcal{L}_{\mathrm{CE}}+\alpha \mathcal{L}_{\mathrm{KD}},
\end{equation}
where $\alpha$ is a weight to balance two losses.

\section{Methodology}

The core idea of our work is to enhance the NMT model with a nearest neighbor retrieval mechanism in a training manner, and thus quantitatively evaluated, the model can perform as well or better than vanilla $k$NN-NMT without any additional decoding cost. In Section~\ref{sec-knnkd}, we first introduce $k$ Nearest Neighbor Knowledge Distillation ($k$NN-KD) to distill the knowledge of $k$NN into a base NMT model. And then, we provide the theoretical analysis in Section~\ref{sec-theory} to support that our method can help to address the overcorrection issue.

\subsection{Nearest Neighbor Knowledge Distillation}
\label{sec-knnkd}

When we apply vanilla $k$NN-MT for testing using beam search with $B$, the time complexity of it is $\mathcal{O}(|\mathcal{D}| B n)$. Compared with the standard beam search whose time complexity is $\mathcal{O}(|\mathcal{V}| B n)$, the decoding speed of vanilla $k$NN-MT is prohibitively slow. This is mainly because vanilla $k$NN-MT has to conduct a $k$NN search over an extremely large datastore $\mathcal{D}$ for each decoding step of each beam.  We propose to move this time-consuming search process forward to the preprocessing phase which can be done offline before training and inference. As shown in Figure~\ref{fig-knnkd}, our proposed $k$NN-KD can be described as follows:

\noindent{\textbf{Step 1: Datastore Building}
} \quad We build the datastore $\mathcal{D}$ in the same way as vanilla $k$NN-MT~\cite{knnmt} which has been described in Section~\ref{sec-vanilla-knnmt}, so we omit it here.

\noindent{\textbf{Step 2: $k$NN Search in Advance} 
} \quad For all the translation contexts $(\mathbf{x}, \mathbf{y}^\star_{<i})$ in the training set, we conduct a $k$NN search using $f\left(\mathbf{x}, \mathbf{y}^\star_{<i}\right)$ as a query to search through the datastore $\mathcal{D}$ built in Step 1, and then we obtain the retrieved results $\mathcal{N}^i = \left\{\left(\mathbf{k}_{j}, v_{j}\right), j \in\{1,2, \ldots, k\}\right\} $. Note that we are performing $k$NN search for \textbf{training set} translation contexts on the datastore built with the \textbf{training set}, which is equivalent to extending the training data by adding $k$ reasonable target tokens for every translation context. Formally, by conducting $k$NN search in advance, we extend the target sentence in the  training set from $\mathbf{y}^\star=\left\{y^\star_{1}, \cdots, y^\star_{n}\right\}$ to $\mathbf{y}^\star=\left\{ \left(y^\star_{1}, \mathcal{K}^1 \right), \cdots, \left(y^\star_{n}, \mathcal{K}^n \right)\right\}$, where $\mathcal{K}^i = \left\{\left(d\left(\mathbf{k}_{j}, f\left(\mathbf{x}, \mathbf{y}^\star_{<i}\right) \right), v_{j}\right), j \in\{1,2, \ldots, k\}\right\}  $.

\noindent{\textbf{Step 3: $k$NN as a Teacher} 
} \quad In the training phase, a $k$NN distribution can be formulated as:
\begin{equation}
\begin{aligned}
\label{eq-knn-teacher-prob}
p^{\mathrm{T}}_{\mathrm{kNN}} & \left( y_{i} \mid \mathbf{x}, \mathbf{y}^\star_{<i}\right) \propto \\
&\sum_{\left(d_{j}, v_{j}\right) \in \mathcal{K}^i } \mathbbm{1}_{y_{i}=v_{j}} \exp \left(\frac{-d_j}{\tau} \right),
\end{aligned}
\end{equation}

We then use $p^{\mathrm{T}}_{\mathrm{kNN}}$ as a teacher to train the base NMT model, and the knowledge distillation objective in Equation~\ref{eq-single-kd} can be rewritten as:
\begin{equation}
\begin{aligned}
\label{eq-knn-kd}
\mathcal{L}_{\mathrm{kNN-KD}}=&-\sum_{y_i\in \mathcal{V}} p^{\mathrm{T}}_{\mathrm{kNN}} \left( y_{i} \mid \mathbf{x}, \mathbf{y}^\star_{<i} \right) \\
& \times \log p_{\mathrm{MT}}\left(y_i \mid \mathbf{x}, \mathbf{y}^\star_{<i}\right).
\end{aligned}
\end{equation}
And the final training objective in Equation~\ref{eq-kd-final} can be rewritten as:
\begin{equation}
\label{eq-knn-kd-final}
\mathcal{L}=(1- \alpha)\mathcal{L}_{\mathrm{CE}}+\alpha \mathcal{L}_{\mathrm{kNN-KD}},
\end{equation}
where $\mathcal{L}_{\mathrm{CE}}$ can be calculated as Equation~\ref{eq-ce}.

\noindent{\textbf{Step 4: Decoding} 
} \quad During inference, our model remains in the standard Seq2Seq manner \cite{transformer}, so we omit it here. 
\subsection{Theoretical Analysis}
\label{sec-theory}
In this section, we show that our proposed $k$NN-KD can help address the overcorrection issue from the perspective of gradients. The gradient of the final objective in Equation~\ref{eq-knn-kd-final} with respect to the logit $\mathbf{z}_{y_i}, {y_i} \in \mathcal{V}$ is:
\begin{equation}
\small
\begin{aligned}
\label{eq-gradient}
&\frac{\partial \mathcal{L}}{\partial \mathbf{z}_{y_i}} =(1-\alpha) \left(p(y_i) - \mathbbm{1}_{{y_i}=y_{i}^{*}} \right) + \alpha \left(p(y_i) - p^{\mathrm{T}} ( y_i) \right) \\
&=\begin{cases}
			p(y_i) -\alpha p^{\mathrm{T}} (y_i), & \text{if } y_i\neq y^\star_i \text{ and } y_i \in \mathcal{K}^i\\
			p(y_i), & \text{if } y_i\neq y^\star_i \text{ and } y_i \notin \mathcal{K}^i \\
			p(y_i) - \left(1 - \alpha + \alpha p^{\mathrm{T}} (y_i) \right), & \text{if } y_i=y^\star_i  
		\end{cases}
\end{aligned}
\end{equation}
where we abbreviate $p_{\mathrm{MT}}\left(y_i \mid \mathbf{x}, \mathbf{y}^\star_{<i}\right)$ to $p(y_i)$ and  $p^{\mathrm{T}}_{\mathrm{kNN}} \left( y_{i} \mid \mathbf{x}, \mathbf{y}^\star_{<i} \right)$ to $p^{\mathrm{T}} (y_i)$. 

For every gradient update in the training phase, the model is trained to decrease the gradient norm to $0$ to reach a local minimum \cite{sg-lm}. Therefore, for the tokens that are reasonable but not ground-truth (i.e., $y_i\neq y^\star_i \text{ and } y_i \in \mathcal{K}^i$), the model has to learn to increase the probability $p(y_i)$ by the degree of $\alpha p^{\mathrm{T}} (y_i)$ so that the gradient norm $|p(y_i) -\alpha p^{\mathrm{T}} (y_i)|$ can reach $0$. For the other non-ground-truth token (i.e., $y_i\neq y^\star_i \text{ and } y_i \notin \mathcal{K}^i$), $p^{\mathrm{T}} (y_i)$ is equal to $0$ since $y_i$ is not included in the retrieved results of $k$NN search, and the model will learn to assign much lower probability $p(y_i)$ to reduce $|p(y_i)|$. Besides, since we build the datastore and conduct $k$NN search on the same training set data, for any translation context, its nearest neighbor over the datastore must be itself, which means if $y_i=y^\star_i $, then $y_i \in \mathcal{K}^i$. Then, for the ground-truth token (i.e.,  $y_i=y^\star_i$), the model is trained to increase the probability $p(y_i)$ by the degree of $\left(1 - \alpha + \alpha p^{\mathrm{T}} (y_i) \right)$. Note that, the gradient norm of the standard CE loss is $|p(y_i) - 1|$ for $y_i=y^\star_i$, and thus that standard CE increases the probability $p(y_i)$ by the degree of $1$. This demonstrates that our $k$NN-KD still makes the model learn to predict the ground-truth but with a relatively lower strength than the standard CE.

Taking the case in Figure~\ref{fig-knnkd} as an example, given the translation context \textit{``Vielen Dank für Ihren hilfreichen Vorschlag || Thanks for your''}, its ground-truth target token is \textit{``useful''}, while \textit{``helpful''} is also reasonable for this translation. Assuming that we have conducted the $k$NN search with $k=3$ in advance as shown in Figure~\ref{fig-knnkd}, and set $\tau$ to $1$, we can then compute the $k$NN teacher distribution as:
\begin{equation}
    p^T(y_4)=\begin{cases}
			0.378, & \text{if } y_4  \text{ is \textit{``useful''} } \\
			0.622, & \text{if } y_4  \text{ is \textit{``helpful''} } \\
			0, & \text{ otherwise } 
		\end{cases}
\end{equation}

According to Equation~\ref{eq-gradient}, the gradient norms are $\left|p(\text{\textit{``helpful''}}) - 0.622\alpha\right|$ for \textit{``helpful''}, and $\left|p(\text{\textit{``useful''}}) - (1 - 0.622\alpha)\right|$ for \textit{``useful''}. Therefore, our $k$NN-KD can train the model to learn from the $k$NN model to increase the probability of \textit{``helpful''} that is reasonable but not ground-truth, thus addressing the overcorrection issue.
\section{Experiments}
\label{sec-experiments}
\subsection{Datasets}
We conduct experiments on IWSLT'14 German-English (De-En, $160k$ training samples), IWSLT'15 English-Vietnamese (En-Vi, $113k$ training samples), and multi-domains translation datasets \cite{multi-domain-data}  (De-En,  $733k$ training samples). For IWSLT'14 De-En, we follow the preprocessing steps provided by fairseq\footnote{\url{https://github.com/pytorch/fairseq/blob/main/examples/translation/prepare-iwslt14.sh}}~\cite{fairseq} to split the data, and process the text into bytepair encoding (BPE) \cite{bpe}. For IWSLT'15 En-Vi, we use the pre-processed dataset\footnote{\url{https://nlp.stanford.edu/projects/nmt/}} provided by \citet{StanfordNMT}. We use tst2012 as the validation set and tst2013 as the test set, which contains $1,553$ and $1,268$ sentences respectively. For multi-domains translation datasets, we use the pre-processed dataset\footnote{\url{https://github.com/zhengxxn/adaptive-knn-mt}} provided by \citet{adaptive-knnmt}, and consider domains including \textit{Koran}, \textit{Medical}, and \textit{Law} in our experiments.

\subsection{Competitive Models}
The proposed $k$NN-KD is an architecture-free method that can be applied to arbitrary Seq2Seq models, which is orthogonality to previous works that design delicate structures to improve performance. Therefore, we mainly compare $k$NN-KD with vanilla $k$NN-MT and some typical KD methods:
\begin{itemize}
    \item \textbf{Word-KD}~\cite{KD}. As described in Section~\ref{sec-kd}, Word-KD is the standard KD that distills knowledge equally for each word.
    \item \textbf{Seq-KD}~\cite{SeqKD}. In this method, teacher model T first generates an extra dataset by running beam search and taking the highest-scoring sequence. Then student model S is trained on this teacher-generated data, and the training objective can be formulated as:
    \begin{equation}
    \begin{aligned}
    \mathcal{L}_{\mathrm{Seq-KD}}=-\sum_{i=1}^{n} \sum_{y_i \in \mathcal{V}} \mathbbm{1}_{y_{i}= \hat{y}_i} \\
    \times \log p_{\mathrm{MT}}\left(y_{i} \mid \mathbf{x}, \hat{\mathbf{y}}_{<j} \right),
    \end{aligned}
    \end{equation}
    where $\hat{\mathbf{y}}$ is the target sequence generated by teacher model, and $n$ is the length of it.
    \item \textbf{BERT-KD}~\cite{BERT-KD}. This method distills knowledge learned in BERT~\cite{BERT} to the student NMT model to improve translation quality.
    \item \textbf{Selective-KD}~\cite{SelectiveKD}. This work finds that some of the teacher's knowledge will hurt the effect of KD, and then address this issue by introducing Selective-KD to select suitable samples for distillation.
\end{itemize}

\begin{table}[t]
\begin{center}
\begin{tabular}{l|ccc}
\toprule
Datasets       & $|\mathcal{D}|$        & $k$  & $\tau$ \\ \hline
IWSLT'14 De-En & 3,949,106  & 64 & 100 \\
IWSLT'15 En-Vi & 3,581,500  & 64 & 100 \\
Koran          & 524,374    & 16 & 100 \\
Medical        & 6,903,141  & 4  & 10  \\
Law            & 19,062,738 & 4  & 10  \\
\bottomrule
\end{tabular}
\end{center}
\caption{\label{tab-preprocess-setting}   Hyper-parameter settings for different datasets. }
	\vskip -0.1in
\end{table}

\subsection{Implementation Details}
All the algorithms are implemented in Pytorch with fairseq toolkit~\cite{fairseq}, and all the experiments are conducted on a machine with 8 NVIDIA GTX 1080Ti GPUs. 
Other details of the experimental setup can be seen in Appendix~\ref{appendix-detailsetup}. 

\noindent{\textbf{Model Configuration} 
} \quad  We choose Transformer~\cite{transformer} as our base NMT model. For IWSLT'14 De-En and IWSLT'15 En-Vi, we use $transformer\_iwslt\_de\_en$ configuration, which has $6$ layers in both encoder and decoder, embedding size $512$,
feed-forward size $1,024$ and attention heads $4$. For  multi-domains translation datasets, we follow \citet{knnmt} to adopt $transformer\_wmt19\_de\_en$ configuration, which has $6$ layers in both encoder and decoder, embedding size $1,024$,
feed-forward size $8,192$ and attention heads $8$.

\begin{table*}[t]
\begin{tabular}{l|ccc|ccc}
\toprule
\multirow{2}{*}{Models}       & \multicolumn{3}{c|}{De-En}                                                            & \multicolumn{3}{c}{En-Vi}                                                             \\
             & BLEU           & upd/s                & token/s                  & BLEU           & upd/s                &  token/s                   \\ \hline
Transformer  & 34.11          & 2.02(1.00$\times$) & 3148.10(1.00$\times$) & 30.76          & 2.55(1.00$\times$) & 2870.07(1.00$\times$) \\
Word-KD      & 34.26          & 1.77(0.88$\times$) & 3291.28(1.06$\times$) & 30.98          & 2.14(0.84$\times$) & 2782.53(0.97$\times$) \\
Seq-KD       & 34.60          & 2.14(1.06$\times$) & 3409.86(1.08$\times$) & 31.20          & 2.80(1.10$\times$) & 2855.77(1.00$\times$) \\
BERT-KD      & 35.63          & 1.70(0.84$\times$) & 3275.43(1.04$\times$) & 31.51          & 2.14(0.84$\times$) & 2785.69(0.97$\times$) \\
Selective-KD & 34.38          & 1.72(0.85$\times$) & 3365.68(1.07$\times$) & 31.48          & 2.09(0.82$\times$) & 3044.68(1.06$\times$) \\ 
$k$NN-MT       & 36.17          & -                               & 920.72(0.29$\times$)  & 32.08          & -                               & 617.88(0.22$\times$)  \\\hline
$k$NN-KD       & \textbf{36.30} & 2.14(1.06$\times$) & 3321.24(1.05$\times$) & \textbf{32.27} & 2.60(1.02$\times$) & 2879.68(1.00$\times$) \\
\bottomrule
\end{tabular}
\caption{\label{tab-iwlst-result} Experimental results on IWSLT'14 De-En and IWSLT'15 En-Vi translation tasks. ``-'' means ``not applicable'' since vanilla $k$NN-MT can only be adopted in the decoding phase. ``upd/s'' and  ``token/s'' are abbreviated notations for ``training updates per second'' and ``generated tokens per second''. }
\end{table*}
\begin{table*}[t]
\resizebox{\textwidth}{!}{
\begin{tabular}{l|cc|cc|cc}
\toprule
\multicolumn{1}{c|}{\multirow{2}{*}{Models}} & \multicolumn{2}{c|}{Koran} & \multicolumn{2}{c|}{Medical} & \multicolumn{2}{c}{Law} \\
\multicolumn{1}{c|}{}                        & BLEU        & token/s      & BLEU         & token/s       & BLEU      & token/s     \\ \hline
Pre-trained Model                      & 16.26       & 1038.97(1.00$\times$)      & 39.91        & 1765.56(1.00$\times$)       & 45.71     & 2404.31(1.00$\times$)     \\
$k$NN-MT                                       & 19.45       & 246.17(0.24$\times$)       & 54.35        & 701.29(0.40$\times$)       & 61.78     & 853.66(0.36$\times$)      \\
Transformer                                  & 13.84       & 1297.45(1.25$\times$)    & 27.51        & 1073.53(0.61$\times$)     & 60.77     & 1689.89(0.70$\times$)       \\  \hline
$k$NN-KD                                       & \textbf{24.86}       & 1236.23(1.19$\times$)    &  \textbf{56.50}         & 1853.58(1.05$\times$)         &  \textbf{61.89}     & 2456.62(1.02$\times$) \\
\bottomrule
\end{tabular}}
\caption{\label{tab-multi-domains-result} Experimental results on multi-domains translation datasets. We leave out the metric for training efficiency (upd/s) since it is only applicable for Transformer and $k$NN-KD, and the training efficiency of these two models are basically the same. }
\end{table*}

\noindent{\textbf{Preprocessing Details} 
} \quad When building the datastores, we use the context vectors input to the final output layer as keys in the datastore $\mathcal{D}$. For IWSLT datasets, the base NMT model is used to obtain the context vectors, while for multi-domains translation datasets, we follow \citet{knnmt} to build datastores by the pre-trained model\footnote{\url{https://github.com/pytorch/fairseq/tree/main/examples/wmt19}}. According to the model configuration, the keys are $512$-dimensional and  $1024$-dimensional for IWSLT datasets  and multi-domains translation datasets, respectively. We use FAISS~\cite{FAISS} for the nearest neighbor search. And we conduct grid searches over $k \in \{4, 8, 16, 32, 64, 128, 256, 512, 1024 \}$ and $\tau \in \{1, 10, 50, 100, 200, 500, 1000\}$, and choose the final settings according to the best BLEU score on the validation set. The final hyper-parameter settings are shown in Table~\ref{tab-preprocess-setting}.

\noindent{\textbf{Evaluation} 
} \quad For all the datasets, we use the beam search with beam size $5$. We evaluate the translation in terms of quality and efficiency. 
\begin{itemize}
    \item \textbf{Quality.} For IWSLT'14 De-En and IWSLT'15 En-Vi, following the common practice, we measure case sensitive BLEU by \textit{multi-bleu.perl}\footnote{\url{https://github.com/moses-smt/mosesdecoder/blob/master/scripts/generic/multi-bleu.perl}}. For  multi-domains translation datasets, we closely follow \citet{knnmt} to evaluate the results by SacreBLEU~\cite{sacre-bleu} for a fair comparison.
    \item \textbf{Efficiency.} We evaluate the efficiency of training and inference by the training updates per second (upd/s) and the generated tokens per second (token/s), respectively.
\end{itemize}

\subsection{Main Results}
\noindent{\textbf{Results of IWSLT Datasets} 
} \quad We first compare $k$NN-KD with vanilla $k$NN-MT and other KD methods on the two IWSLT translation tasks.  Note that there are several hyper-parameters in vanilla $k$NN-MT: tunable weight ($\lambda$), number of neighbors per query ($k$), and temperature ($\tau$). These hyper-parameters have great effects on the translation results. We also conduct grid searches over these hyper-parameters, and find the best settings according to BLEU score on the validation set.

As shown in Table~\ref{tab-iwlst-result}, $k$NN-KD outperforms all the other strong baselines on both IWSLT datasets, e.g., an improvement of $+2.14$ and $+1.51$ BLEU score over Transformer. Moreover, we observe that our proposed $k$NN-KD can even perform better than vanilla $k$NN-MT, while gaining a significant speedup. On the one hand, $k$NN-KD, like other KD methods, maintains the standard Seq2Seq manner at inference time, thus keeping the same decoding speed as Transformer. On the other hand, $k$NN-KD also keeps the same training speed as Transformer, and it is more efficient than Word-KD, BERT-KD and Selective-KD. This is because the calculation of the teacher model distribution $p^{\mathrm{T}}_{\mathrm{kNN}} \left( y_{i} \mid \mathbf{x}, \mathbf{y}^\star_{<i} \right)$ only needs to be performed on a relatively small $k$NN retrieved set $\mathcal{K}^i$, while word-level KD have to compute the teacher distribution over the whole vocabulary $\mathcal{V}$. 

\begin{table}[t]
\resizebox{\columnwidth}{!}{
\begin{tabular}{l|cc}
\toprule
Models      & Law$\rightarrow$Medical & Medical$\rightarrow$Law \\\hline
Transformer & 18.73                     & 2.07  \\
$k$NN-KD      & 22.31                     & 14.82         \\
\bottomrule
\end{tabular}}
\caption{\label{tab-generality-result} Generalizability Evaluation. ``Law$\rightarrow$Medical'' means that we train the model on the Law domain and directly apply it to Medical domain, and vice versa. The results are BLEU scores.}
\end{table}

\noindent{\textbf{Results of Multi-domains Datasets} 
} \quad Apart from IWSLT datasets, we further compare our $k$NN-KD with $k$NN-MT on multi-domains translation datasets. First, we follow \citet{knnmt} to conduct inference with the pre-trained model and vanilla $k$NN-MT. Then, we train the base NMT model using standard CE and $k$NN-KD on each domain's training data, and report the results  in Table~\ref{tab-multi-domains-result} as a comparison. In all domains, $k$NN-KD again outperforms all the baselines. Most importantly, our proposed $k$NN-KD can achieve a consistent improvement over vanilla $k$NN-MT ($+2.56$ BLEU score on average) with a significant speedup. This further confirms the effectiveness and efficiency of our method.

\noindent{\textbf{Generalizability} 
} \quad To verify the generalizability of our method, we further conduct experiments on the scenario that we train a NMT model on a specific domain and evaluate it on the out-of-domain test set. As shown in Table~\ref{tab-generality-result}, our $k$NN-KD performs significantly better than Transformer trained by standard CE. It proves the statement in Section~\ref{sec-introduction} that compared with standard CE, $k$NN-KD can improve the generalizability of NMT models.

\subsection{Analysis}
There are two key hyper-parameters in our $k$NN-KD: number of neighbors per query ($k$), and temperature ($\tau$). In this section, we investigate the effects of these two hyper-parameters on the validation set of IWSLT'14 De-En.

\begin{figure}[t]
	\begin{center}
		\centerline{\includegraphics[width=\columnwidth]{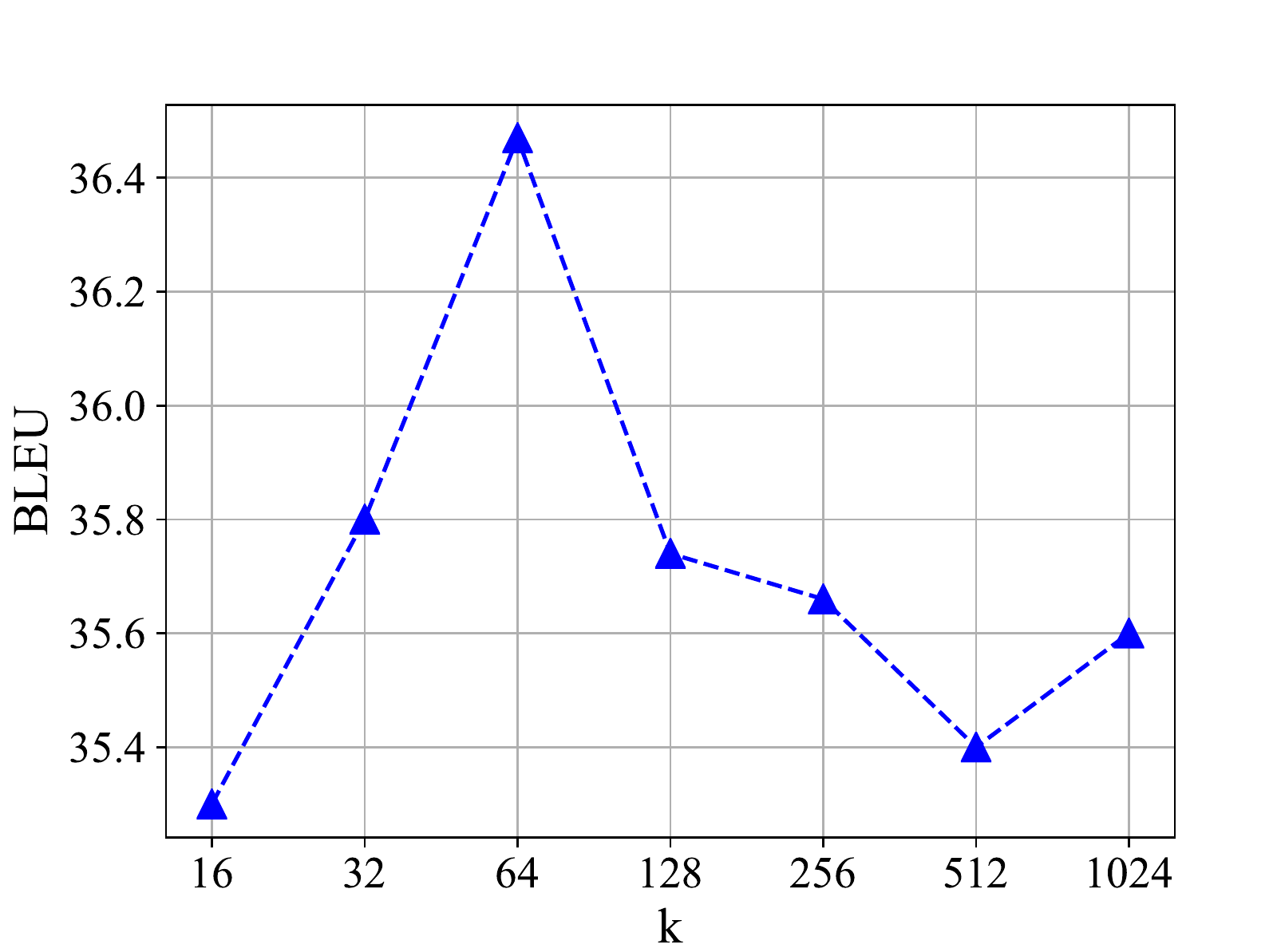}}
		\caption{BLEU scores with different $k$ and fixed $\tau$ ($\tau = 100$) on the validation set of IWSLT'14 De-En dataset.}
		\label{fig-effect-k}
	\end{center}
	\vskip -0.2in
\end{figure}

\noindent{\textbf{Effect of $k$} 
} \quad We fix the temperature $\tau$ to $100$, and train the model using $k$NN-KD with different $k$. As shown in Figure~\ref{fig-effect-k}, the BLEU score first rises with the increase of $k$, and reaches the best performance peak when $k=64$. And then, performance deteriorates with a larger $k$. This suggests that, the retrieved results of $k$NN search can substantially improve training when $k$ is relatively small, but it will also introduce more noise when $k$ gets larger.

\begin{figure}[t]
	\begin{center}
		\centerline{\includegraphics[width=\columnwidth]{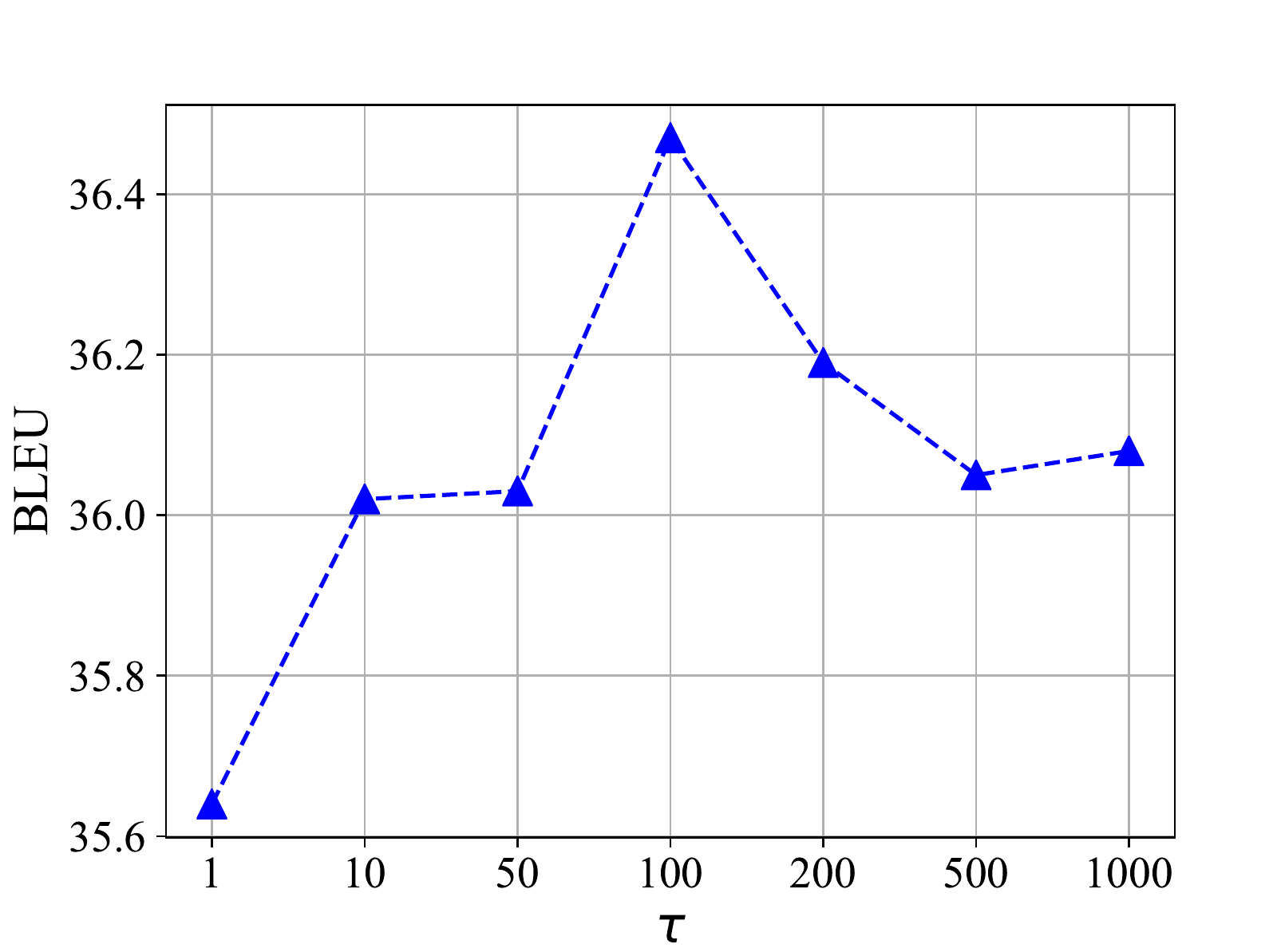}}
		\caption{BLEU scores with different $\tau$ and fixed $k$ ($k=64$) on the validation set of IWSLT'14 De-En dataset.}
		\label{fig-effect-tau}
	\end{center}
	\vskip -0.2in
\end{figure}

\noindent{\textbf{Effect of $\tau$} 
} \quad We train the model using $k$NN-KD with different $\tau$ and fixed $k$ ($k=64$). As shown in Figure~\ref{fig-effect-tau}, a temperature of $1$ results in a significantly lower BLEU score than those greater than $1$. This is because a large temperature value can flatten the $k$NN teacher distribution in Equation~\ref{eq-knn-teacher-prob} to prevent assigning most of the probability mass to a single neighbor. The results show that for $k=64$, the optimal temperature is $100$.


\begin{figure}[t]
	\begin{center}
		\centerline{\includegraphics[width=\columnwidth]{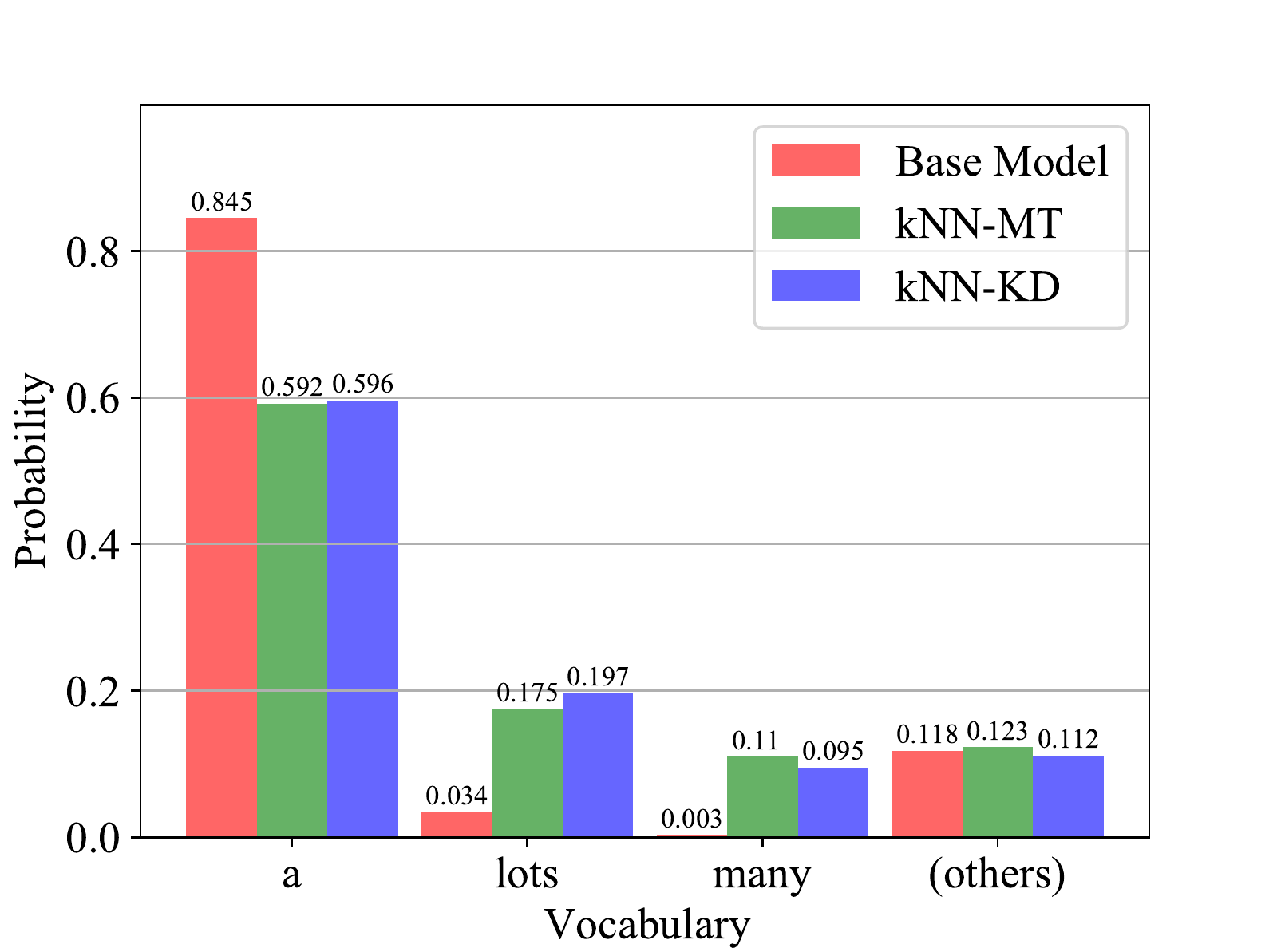}}
		\caption{Predicted probabilities output from the base NMT model, $k$NN-MT and our $k$NN-KD, given the translation context \textit{``es gibt eine menge geschichten darüber , warum wir dies getan haben . || there are''} }
		\label{fig-predicted-probs}
	\end{center}
	\vskip -0.2in
\end{figure}

\subsection{Case Study} 
In this section, we show how our proposed method works by presenting a real case. There exists an example in the test set of IWSLT'14 De-En that the source sentence is \textit{``es gibt eine menge geschichten darüber , warum wir dies getan haben .''} and the corresponding target sentence is \textit{``there are a lot of stories about why we did this .''}. Given the source sentence and target subsequence \textit{``there are''} as the translation context, \textit{``many...''}, \textit{``lots of...''}, and \textit{``a lot of...''} are all correct translations. We input this translation context to the base NMT model, $k$NN-MT, and our model, and observe the predicted probabilities over the vocabulary. As shown in Figure~\ref{fig-predicted-probs}, all the models predict \textit{``a''} with the maximal probability that matches the ground-truth. However, since the base model is trained by CE loss using one-hot vector as supervision, it suffers from a serious overcorrection problem that the model assigns an overconfident probability to the token \textit{``a''} and almost none to other reasonable target tokens such as \textit{``lots''} and \textit{``many''}. On the contrary, both $k$NN-MT and our $k$NN-KD increase the probabilities of the reasonable target tokens, and these two models have similar predicted probabilities. Note that $k$NN-MT obtains this probability distribution by interpolating the base NMT probability with a $k$NN search probability at decoding time, while our $k$NN-KD directly outputs this distribution without any additional operations. This further confirms that $k$NN-KD can train the model to learn the knowledge of $k$NN that prevents the over-confidence of the model on the one-hot label, thus leading to the better generalizability for inference.

\section{Related Works}
\subsection{Neural Machine Translation}

Machine translation has developed rapidly in recent years. The early models were mainly based on statistical machine learning \cite{brown1990statistical, och2003minimum, koehn2007moses}. Then, with the development of deep learning technology, many models used RNN\cite{sutskever2014sequence, bahdanau2014neural}, CNN\cite{gehring2017convolutional}, or Transformer\cite{transformer} as their backbones.

Recently, a few studies have combined $k$ nearest neighbors algorithm closely with NMT models to improve performance. \citet{knnmt} used a nearest neighbor classifier to predict tokens on a large datastore of cached examples and proposed $k$NN-MT. However, \citet{fast-knnmt} pointed out that $k$NN-MT is two-order slower than vanilla MT models, which limits the deployment for real-world applications. They proposed Fast $k$NN-MT to solve this problem. \citet{faster-knnmt} also noticed the low-efficiency problem of $k$NN-MT. Thus, they used a hierarchical clustering strategy and proposed Faster $k$NN-MT.
Although the above studies have made feasible fixes, $k$NN search is still required in the decoding phase, which dramatically increases the difficulty of practical applications compared to standard MT models.

\subsection{Knowledge Distillation}

Knowledge distillation (KD) introduces teacher network and student network to help knowledge transfer and it was widely used in NMT \cite{hinton2015distilling}. \citet{SeqKD} introduced two sequence-level KD methods to improve the performance of NMT. \citet{miceli2017regularization} used KD to address the problem of catastrophic forgetting in the fine-tuning stage. \citet{tan2019multilingual} used KD to enable the multilingual model to fit the training data and to match the outputs of the teacher models. \citet{clark2019bam} distilled single-task models into one multi-task model. \citet{BERT-KD} used BERT as the teacher model after fine-tuning on the target generation tasks to improve the conventional Seq2Seq models. \citet{SelectiveKD} proposed batch-level and global-level selection strategies to choose appropriate samples for knowledge distillation. We focus
on using KD to leverage the knowledge retrieved by $k$NN search to enhance a base NMT model.

\section{Conclusion}
In this paper, we introduce $k$NN-KD that distills the knowledge retrieved by $k$NN search to prevent the base NMT model from overcorrection.  Experiments show that $k$NN-KD can improve over vanilla $k$NN-MT and other baselines without any additional cost for training and decoding. In the future, we will apply $k$NN-KD to many other tasks. We will also explore the effect of $k$NN-KD on improving the diversity of text
generation.

\section*{Acknowledgements}
This work was supported by National Science Foundation of China (No. 62161160339), National Key R\&D Program of China (No.2018YFB1005100), State Key Laboratory of Media Convergence Production Technology and Systems and Key Laboratory of Science, Techonology and Standard in Press Industry (Key Laboratory of Intelligent Press Media Technology). We appreciate the anonymous reviewers for their helpful comments. Xiaojun Wan is the corresponding auhor.

\bibliography{anthology,custom}
\bibliographystyle{acl_natbib}

\appendix


\section{Experimental Setup}
\label{appendix-detailsetup}
\subsection{Datasets}

The dataset statistics for all the datasets are reported in Table~\ref{tab-samples-num}. It is worth to mention that IWSLT datasets are under the Creative Commons BY-NC-ND license, and the multi-domains translation datasets are under the BSD license.

\begin{table}[h]
\begin{center}
\begin{tabular}{c|ccc}
\toprule
           & Train       & Valid   & Test    \\ \hline
IWLST‘14 De-En & 160,239      & 7,283      & 6,750      \\
IWLST‘15 En-Vi & 133,166      & 1,553    & 1,268      \\
Koran & 17,982     & 2,000     & 2,000      \\
Medical & 248,099      & 2,000     & 2,000      \\
Law & 467,309     & 2,000     & 2,000       \\
\bottomrule
\end{tabular}
\end{center}
\caption{The number of examples for different datasets.}
\label{tab-samples-num}
\end{table}

\subsection{Hyper-parameters Setting}

All the algorithms are implemented in Pytorch with fairseq toolkit~\cite{fairseq}, and all the experiments are conducted on a machine with 8 NVIDIA GTX 1080Ti GPUs with the hyperparameters reported in Table~\ref{Hyperparameters}.
\begin{table}[h]
\begin{center}
\begin{tabular}{l |c | c}
\toprule
 Hyperparameters & IWSLT & Multi-domains \\
\hline 
 Max tokens & 8192 & 1280 \\
 Learning rate & 5e-4 & 5e-4 \\
 LR scheduler & Inverse sqrt & Inverse sqrt \\
 Minimal LR & 1e-9 & 1e-9 \\
 Warm-up LR  & 1e-7 & 1e-7 \\
 Warm-up steps & 4000 & 4000 \\
 Gradient clipping & 0.0 & 0.0 \\
 Weight decay & 0.0 & 0.0001\\
 Droupout & 0.3 & 0.2 \\
 Attention dropout & 0.0 & 0.1 \\
 Activation dropout & 0.0 & 0.1 \\
 $\alpha$ in Equation~\ref{eq-knn-kd-final} & 0.5 & 0.5 \\
 Optimizer  & \text { Adam } & \text { Adam }\\
\quad  -$\beta_{1}$& 0.9 & 0.9  \\
\quad  -$\beta_{2}$  & 0.98 & 0.98 \\
\quad  -$\epsilon$ & 1$\mathrm{e}$-8 & 1$\mathrm{e}$-8\\
\bottomrule
\end{tabular}
\end{center}
\caption{Hyperparameter settings for different datasets.}
\label{Hyperparameters}
\end{table}

Note that during training, we are using the dynamic batching provided by fairseq, and choose the max tokens according to the GPU memory constraint. We train the model for $200$ epochs on IWSLT datasets, $250$ epochs on Koran domain, $100$ epochs on Medical domain, $120$ epochs on Law domain, while the early-stop mechanism is also adopted with patience set to $20$.

\section{Limitation and Potential Risks}

Although $k$NN-KD is efficient in both training and inference, it will take a relatively long time for preprocessing to build the datastore and conduct $k$NN searches, and it also requires large disk space to store all these results. However, since the preprocessing can be done offline, it does not limit the deployment of $k$NN-KD in real-world applications. 

Our model is trained on open source datasets, and thus if there exists toxic text in the training data, our model may have the risk of producing toxic content.

\end{document}